\documentclass[10pt,twocolumn,letterpaper]{article}

\usepackage[pagenumbers]{cvpr} % To force page numbers, e.g. for an arXiv version
\usepackage{algorithm}
\usepackage{algorithmic}

\usepackage{amsmath}

\usepackage{multirow}
\usepackage[accsupp]{axessibility}

\usepackage{enumitem}
\usepackage{graphicx}
\usepackage{amssymb}
\usepackage{booktabs}

\usepackage{adjustbox}
\usepackage{caption}
\usepackage{makecell}
\usepackage{float}
\usepackage{paralist}
\usepackage{color}
\usepackage{threeparttable}
\usepackage{lipsum}
\usepackage{url}

\usepackage{fancyhdr}

\usepackage[dvipsnames]{xcolor}

\definecolor{cvprblue}{rgb}{0.21,0.49,0.74}
\usepackage[pagebackref,breaklinks,colorlinks,citecolor=cvprblue]{hyperref}

\def\paperID{4403} % *** Enter the Paper ID here
\def\confName{CVPR}
\def\confYear{2025}

\title{Optimizing for the Shortest Path in Denoising Diffusion Model}

\author{
    Ping Chen$^{1,2}$, Xingpeng Zhang$^{4}$, Zhaoxiang Liu$^{1,2*}$, Huan Hu$^{1,2}$, Xiang Liu$^{1,2}$,\\
    Kai Wang$^{1,2}$, Min Wang$^{4}$, Yanlin Qian$^{3}$, Shiguo Lian$^{1,2}$\thanks{Corresponding author.} \\[2mm]
    $^{1}$ Data Science \& Artificial Intelligence Research Institute, China Unicom \\
    $^{2}$ Unicom Data Intelligence, China Unicom,
    $^{3}$ DJI Technology Co.,Ltd. \\
    $^{4}$ School of Computer Science and Software Engineering, Southwest Petroleum University, Chengdu, China \\
    {\tt\small \{chenp181, liuzx178, huh30, liux750, wangk115, liansg\}@chinaunicom.cn}\\ 
    {\tt\small xpzhang@swpu.edu.cn, wangmin80616@163.com, honza.qian@dji.com}
}

\begin{document}

\maketitle

% \begingroup
%   % 修改脚注规则，只在此组内有效：左侧显示一条约2cm长的横线
%   \renewcommand{\footnoterule}{%
%     \noindent\makebox[2cm]{\hrulefill}\vspace{0.5ex}%
%   }
%   % 取消脚注标号
%   \renewcommand{\thefootnote}{}%
%   \footnotetext{\scriptsize $\dagger$~Co-corresponding author \quad $\ddagger$~Corresponding author (Team leader)}
% \endgroup

\begin{abstract}
 %Diffusion model can generate images with high photorealism and has wide application prospects. However, accelerating the generation speed is always the key technology that must be solved due to its internal mechanism. 
 %In this paper, we present a new denoising diffusion method based on shortest path modeling, which aims to improve both the efficiency and quality of the denoising process through optimized residual propagation. 
In this research, we propose a novel denoising diffusion model based on shortest-path modeling that optimizes residual propagation to enhance both denoising efficiency and quality.
 %Our approach builds upon Denoising Diffusion Probabilistic Models (DDPM) and Denoising Diffusion Implicit Models (DDIM), using graph theory to view the denoising process as a shortest path problem that minimizes reconstruction errors. 
 %Based on Denoising Diffusion  Probabilistic Models (DDPM)
Drawing on Denoising Diffusion Implicit Models (DDIM) and insights from graph theory, our model, termed the Shortest Path Diffusion Model (ShortDF), treats the denoising process as a shortest-path problem aimed at minimizing reconstruction error. By optimizing the initial residuals, we improve the efficiency of the reverse diffusion process and the quality of the generated samples.
%Based on Denoising Diffusion Implicit Models (DDIM) and hints from graph theory, our model, termed as Shortest Path Diffusion Model (in short ShortDF) regards the denoising process as the shortest path problem to minimize the reconstruction error. By optimizing the initial residuals, we enhance the efficiency of the reverse diffusion process and improve the quality of generated samples. 
% We codenamed our model the Shortest Path Diffusion Model (ShortDF). 
%While many methods primarily aim to reduce sampling time, our approach achieves faster sampling while also maintaining or even improving generation quality, making it more practical for real-time applications.
% Extensive experiments on several standard benchmarks  prove that our method achieves state-of-the-art performance in both generation speed and sample quality. Our results indicate that this approach reduces diffusion time while maintaining or enhancing the fidelity of generated outputs compared to existing techniques. 
Extensive experiments on multiple standard benchmarks demonstrate that ShortDF significantly reduces diffusion time (or steps) while enhancing the visual fidelity of generated samples compared to prior arts.
% This work opens the door for further study and useful applications in a range of generating tasks by providing a fresh viewpoint and a strong foundation for effective data production using diffusion models.
This work, we suppose, paves the way for interactive diffusion-based applications and establishes a foundation for rapid data generation. Code is available at \href{https://github.com/UnicomAI/ShortDF}{https://github.com/UnicomAI/ShortDF}.
%This work offers a new perspective and a solid framework for efficient data generation using diffusion models, paving the way for future research and practical applications in various generative tasks.
%\vspace{-1ex}
\end{abstract}    
\section{Introduction}
\label{sec:intro}

%Generative models have advanced rapidly in recent years, with Denoising Diffusion Models (DDMs) proving to be highly effective in various tasks, including image generation, speech synthesis, and other generative applications.\cite{Yang2024Survey, Baranchuk2022Segmentation, Brempong2022Denoising, Yang2023Image}. 
Recent years have seen significant advancements in generative models, in which Denoising Diffusion Models demonstrates exceptional performance across various domains, e.g. speech synthesis, 2D or 3D visual assets generation, and other applications \cite{Yang2024Survey, Baranchuk2022Segmentation, Brempong2022Denoising, Yang2023Image}.
These models capture the inherent data distribution by gradually adding random noise to the data and learning the reverse process. 

% Although the quality and diversity of generated outputs have significantly improved, the denoising process continues to be computationally intensive and time-consuming \cite{Yin2023One, Zhou2023Fast}. 
While the quality and diversity of the generation have been massively improved,  concerns still exist on intensive computational load and time consumption \cite{Yin2023One, Zhou2023Fast}.
% These challenges create substantial barriers to the application of diffusion models in real-time scenarios, particularly where quick response times are essential, such as in interactive graphics generation and real-time video processing \cite{Austin2021Discrete, Li2022Text}. 
These burdens the further application of diffusion models in real-time scenarios, especially in cases where quick response time is in need, such as interactive graphic editing and real-time video processing (even we seldom consider)  \cite{Austin2021Discrete, Li2022Text}.
To address these concerns, researchers have been actively investigating more efficient strategies. 
Traditional diffusion models typically require hundreds of iterations to achieve high-quality results, which leads to considerable computational costs \cite{Song2020Improved, Ho2020Denoising}. 
Recent efforts have focused on reducing the complexity through techniques such as knowledge distillation, which packs the multi-step diffusion sampling procedure into a single-step student network \cite{Liu2023Flow, Meng2023Distillation, Yin2023One}. 
However, accurately approximating this high-dimensional and complex sampling process remains a challenging task.

\begin{figure}[t]
    \centering
    \includegraphics[width=\linewidth]{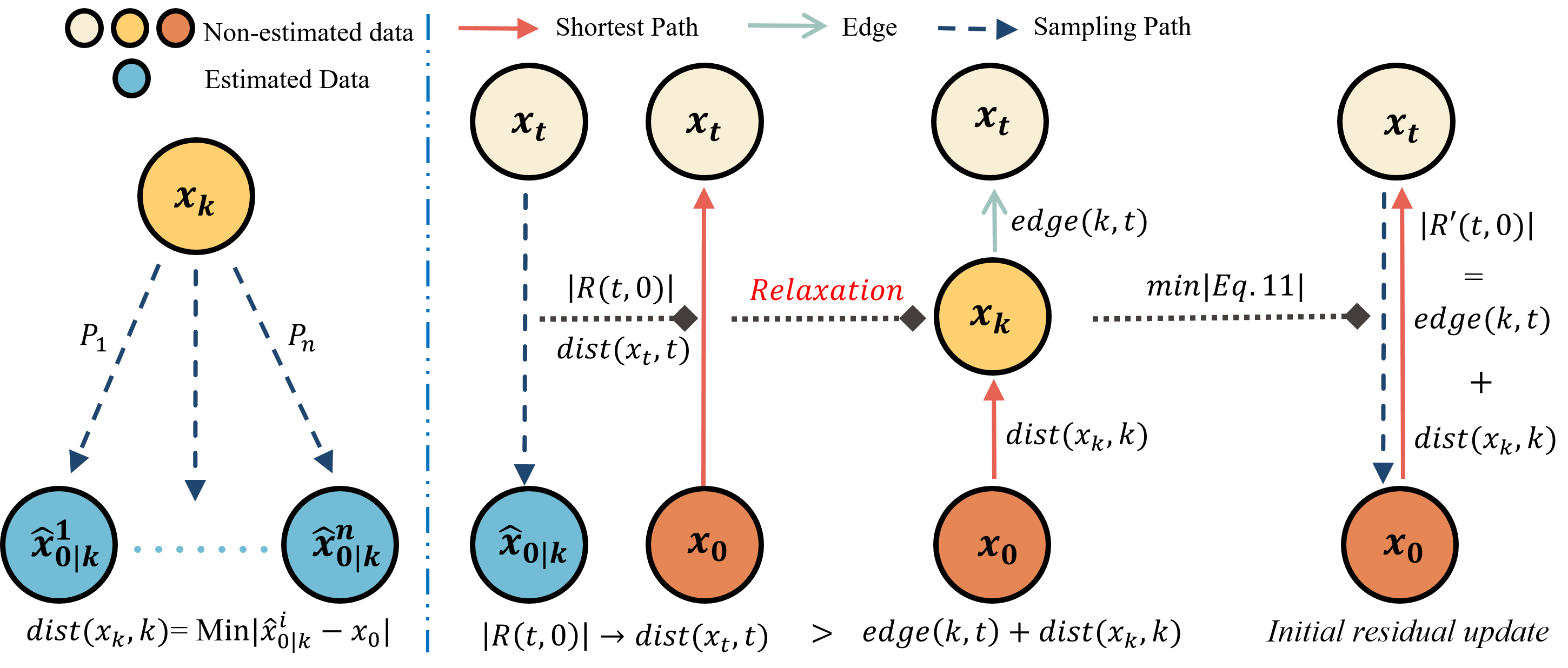}
    \caption{%Modeling each reverse step involves identifying a node that meets the criteria for the shortest path. With abuse of notation,  we bring in a relaxation that can be interpreted as a prior ``A straight line is the shortest path between two points'' and implemented in loss form.  Consequently, we optimized the reverse network to obtain the shortest path.
   % Modeling each reverse step involves identifying a node that meets the criteria for the shortest path. With abuse of notation, we introduce a relaxation that interprets the prior ``a straight line is the shortest path between two points,'' allowing us to treat the initial residual as an alternative to the shortest path. This relaxation is implemented in loss form to guide optimization, enabling the reverse network to iteratively refine the initial residual as though following the shortest path.
   Modeling each reverse step involves identifying a data node $x_t$ that minimizes cumulative transition costs $dist(x_t,t)$ (akin to a shortest path $P_i$ in a weighted graph). By relaxing the strict 'straight-line' analogy to consider minimal-cost paths in the diffusion graph, we treat the initial residual $|R(t,0)|$ as an alternative path candidate. This relaxation is embedded in the loss function, enabling the reverse graph to iteratively refine the residual toward a shorter path (e.g., collapsing $x_0\xrightarrow{}x_k\xrightarrow{}x_t$ into $x_0\xrightarrow{}x_t$), thus optimizing the reconstruction path through dynamic path compression.
    }
    \label{fig_teaser}
\end{figure}

Coupled with distillation-based approaches, other researchers have proposed faster numerical solvers, which rely on the current step size to estimate the next solution \cite{Zhou2023Fast}. 
% Although these methods can decrease the number of diffusion steps from 1000 to fewer than 20, they frequently come with trade-offs in generation quality \cite{Song2021DDIM, Lu2022DPM, Zhang2023Fast}. 
While the number of diffusion steps is decreased from 1000 to fewer than 20, these methods \cite{Song2021DDIM, Lu2022DPM, Zhang2023Fast} are facing a dilemma that generation quality sacrifices.
% Additionally, further reducing the number of sampling steps often results in significant performance degradation\cite{Berthelot2023TRACT}.
On top of that, further reducing the sampling steps often leads to a decline in performance \cite{Berthelot2023TRACT}.
% As a result, achieving faster sampling while maintaining quality has become a key research objective.
In a nutshell, a good trade-off between fast sampling and high generation quality is not easy to achieve, which is we are working on in this paper.

% In this work, we present a novel denoising method that addresses both efficiency and quality through an optimization framework based on shortest-path modeling. 
In this paper, we present ShortDF, which successfully achieves the aforementioned trade-off by utilizing a novel optimization framework that propagates residuals while maintaining high generation quality.
Building on the foundation of Denoising Diffusion Probabilistic Models (DDPM) and Denoising Diffusion Implicit Models (DDIM), we integrate graph-theoretic techniques to reduce reconstruction errors.
By framing the denoising process as an optimization problem for the shortest path in graph theory, we can enhance the efficiency of the procedure without compromising the quality of generation. Fig. \ref{fig_teaser} provides a clear illustration of the concept presented in this paper.

Specifically, Denoising Diffusion Probabilistic Models (DDPM) generate samples by gradually adding Gaussian noise to the original data and then applying reverse denoising. In contrast, Denoising Diffusion Implicit Models (DDIM) accelerate this process by introducing more flexible sampling paths \cite{Ho2020Denoising, Song2021DDIM}.
However, these methods often expose inefficiencies and residual accumulation in managing the multi-step reverse process.
In contrast, our approach presents a novel optimization perspective by framing the denoising process as residual propagation. This method optimizes the initial residuals, thereby enhancing the overall efficiency of the reverse diffusion path.
Till our grasp, we are the first to analyze the dynamics of residual propagation in the denoising process and to construct a reverse-step graph,  enabling optimal transitions between any arbitrary pair of steps.
This graph-based modeling enables us to capture error propagation more effectively. Through iterative relaxation, represented as a loss function based on shortest-path modeling, we gradually reduce the residuals.
Our method accelerates the denoising process and ensures high reconstruction fidelity at each step, thereby overcoming the quality degradation observed in recent fast sampling methods.
Another novelty is that, while perseving the sampler's task-agnostic nature, our method allows end-to-end optimizaiton
of the generator and the sampler and obtaining robustness across domains like text-to-image geneartion.%\cite{Zhou2023Fast, Song2021DDIM}.

% In summary, we present a novel optimization framework for Denoising Diffusion Models that greatly enhances the efficiency of the denoising process while also improving the quality of generated results. 
% Our key innovation lies in designing a loss function based on shortest-path modeling to optimize initial residuals, thereby improving the overall denoising performance. 
% This method is the first to apply shortest path techniques to the design of denoising loss, paving the way for further research and practical applications, particularly in scenarios demanding real-time, high-quality generation.

To summarize the contributions of this work, we highlight the following three points:

{\textbullet} Proposes a novel denoising method (ShortDF) that achieves a balance between fast sampling and high-quality generation through a shortest-path optimization framework.

{\textbullet} Integrates graph-theoretic techniques into the denoising process, reducing reconstruction errors and enhancing efficiency without compromising quality.

{\textbullet} Enables end-to-end optimization of the generator and sampler, and extensive experiments have demonstrated its robustness and effectiveness across various domains.

\section{Related Work}
\label{sec:related}
%Due to its powerful generative capabilities, the diffusion model has achieved remarkable success in image generation, audio synthesis, video generation, and other fields. 
Despite SOTA results,  the inherently iterative procedure of diffusion models entails a high and often prohibitive computational cost for real-time applications \cite{Yin2023One}. The inference process of accelerating diffusion models has been a key focus in the field, and there are currently two main research directions: fast diffusion sampler and diffusion distillation \cite{Yin2023One, Zhou2023Fast}.

\noindent \textbf{Diffusion distillation.} These methods treat diffusion distillation as a form of knowledge distillation \cite{Yin2023One}, where a student model is trained to condense the multi-step outputs of the original diffusion model into a single step. One straightforward approach involves calculating the denoising trajectory in advance and subsequently training the corresponding student model in pixel space \cite{Zheng2023Fast}. However, this method faces the significant challenge of the high computational cost associated with calculating and fitting the complete denoising trajectory. The Progressive Distillation (PD) model \cite{SalimansH2022Progressive, Meng2023Distillation} effectively reduces the number of sampling steps by half. InstaFlow \cite{Liu2023Flow, Liu2024InstaFlow} progressively learns straighter flows, ensuring that the one-step prediction maintains accuracy over a larger distance. Additionally, Consistency Distillation (CD) \cite{Song2023Consistency}, TRACT \cite{Berthelot2023TRACT}, and BOOT \cite{Gu2023boot} aligned the different time steps of the ODE stream with its output, thereby achieving efficient diffusion acceleration. The Variational Score Distillation (VSD) can leverage a pretrained text-to-image diffusion model as a distribution matching loss \cite{Wang2023ProlificDreamer}. Based on VSD, Yin et al. \cite{Yin2023One} proposed a method called distributed matched distillation to generate highly photorealistic images from complex datasets.

\noindent \textbf{Accelerating diffusion sampling.} A substantial body of research on faster diffusion samplers is grounded in solving the probability flow ordinary differential equation (ODE) \cite{Yang2024Survey}. Denoising Diffusion Implicit Models (DDIM) \cite{Song2021DDIM} were among the first initiatives to accelerate sampling in diffusion models, extending the original Denoising Diffusion Probabilistic Model (DDPM) to non-Markovian scenarios. Building on this foundation, Generalized Denoising Diffusion Implicit Models (gDDIM) \cite{Zhang2023gDDIM} introduce a modified parameterization of the scoring network, facilitating deterministic sampling for a broader range of diffusion processes. The Efficient Denoising Model (EDM) \cite{Karras2022EDM} presents a design framework that delineates specific design choices to optimize the diffusion process. Liu et al. proposed Pseudo Numerical Diffusion Models (PNDM) \cite{Liu2022Pseudo}, which employ a numerical solver with a nonlinear transfer component to address a differential equation on manifolds. The Diffusion Exponential Integrator Sampler \cite{Zhang2023Fast} and DPM-Solver \cite{Lu2022DPM} capitalize on the semi-linear structure of the probability flow ODE to create specialized ODE solvers that outperform general-purpose Runge-Kutta methods. It has been observed in the Approximate Mean-Direction Solver (AMED-Solver) \cite{Zhou2023Fast} that nearly every sampling trajectory resides within a two-dimensional subspace of the embedded environment space. This approach eliminates truncation error by directly learning the mean direction, enabling rapid diffusion sampling. The Residual Denoising Diffusion Models (RDDM) \cite{Liu2024Residual} decouple the traditional single denoising diffusion process into residual diffusion and noise diffusion. The residual diffusion component represents directional diffusion from the target image is compared to the degraded input image, explicitly guiding the reverse generation process for image restoration. These numerical methods inherently introduce a degree of approximation error, which can impact the quality of image generation.

\section{Proposed Method}
% In this work, we focus on optimizing the denoising process in diffusion models by formulating it as a shortest-path problem through residual propagation. 
Building on DDPM and DDIM, we employ graph theory methods to minimize reconstruction error by finding the shortest path in the diffusion model.

\subsection{Background: DDPM and DDIM}
\noindent \textbf{Diffusion vs. Inverse Diffusion:} Based on the Markov chain, DDPM progressively adds noise to the clean data $x_0$ in the forward process, resulting in a noisy sample $x_t$ at timestep $t$. In the reverse process, the noise in $x_t$ is gradually eliminated to reveal $x_0$. The forward process is defined as described in \cite{Ho2020Denoising}:
\begin{align}\label{eq-addnoise}
%\small{
    x_t = \sqrt{\bar{\alpha}_t} \cdot x_0 + \sqrt{1 - \bar{\alpha}_t} \cdot \epsilon, \quad \epsilon \sim N(0, \mathbf{I})
%    }
\end{align}
where $t \in [1,T]$, $\bar{\alpha}_t = \prod_{i=1}^t \alpha_i$, and $\alpha_t$ is a predefined variance schedule parameter controlling the strength of noise added at time step $t$. 
In the reverse order, the model estimates noise using a neural network $\epsilon_\theta(x_t, t)$, and the estimated clean sample $\hat{x}_0$ at timestep $t$ is given by:
\begin{align}\label{eq-x0_t}
    \hat{x}_{0|t} = \frac{x_t - \sqrt{1 - \bar{\alpha}_t} \cdot \hat{\epsilon}_t}{\sqrt{\bar{\alpha}_t}}
\end{align}
Where $\hat{\epsilon}_t=\epsilon_{\theta}(x_t, t)$ denotes the network's prediction of the noise. The noise loss:
\begin{align}\label{eq-noiseloss}
    L_\epsilon = \mathbb{E}_{t, x_0, \epsilon} \left[ \| \epsilon - \hat{\epsilon_t} \|_2 \right] 
\end{align}

This loss term ensures that the model accurately predicts the noise, allowing it to effectively denoise the sample over multiple steps throughout the entire reverse path.

\noindent \textbf{Accelerating Sampling with Flexible Paths:}
DDIM accelerates the sampling process by introducing flexibility in the reverse diffusion paths \cite{Song2021DDIM}.
Unlike DDPM, which employs a fully stochastic reverse process, DDIM enables deterministic or partially stochastic sampling, governed by the variance term $\sigma_n$.
The DDIM sampling equation is:
\begin{equation}\label{eq-ddim}  
\begin{aligned}  
\hat{x}_k &= DDIM(\hat{x}_{0|t}, \hat{\epsilon}_t, k, \sigma_n) \\  
    &= \sqrt{\bar{\alpha}_k} \cdot \hat{x}_{0|t} + \sqrt{1 - \bar{\alpha}_k - \sigma_n^2} \cdot \hat{\epsilon}_t + \sigma_n \cdot \epsilon  
\end{aligned}  
\end{equation}
where $k \leq t$. DDIM uses the estimate $\hat{x}_{0|t}$ as a replacement for the clean data $x_0$, reducing the number of reverse diffusion steps. 

% \subsection{Residual Propagation and Path Representation}
\subsection{Residual Propagation as Path Representation}
\label{RPPR}

In DDIM, when $\sigma_n=0$, the reverse process becomes fully deterministic, allowing us to analyze the residuals along the reverse diffusion path. We assume that the sampling path is $k_1\xrightarrow{}\cdots \xrightarrow{}k_n\xrightarrow{}0$, where $k_{i}\in[1,T]$ and $k_{i} \geq k_{i+1}$. 

\noindent \textbf{Residual Propagation:} Let $R(k_i, k_j)$ denote the residual change from the step $k_i$ to $k_j$. 
At timestep $k_1$, the initial residual $R(k_1, 0)$  is the one-step sampling estimation error from $k_1$ to $0$. It quantifies the difference between $x_0$ and its direct estimate $\hat{x}_{0|k_1}$. 
Refering to Eq. \eqref{eq-addnoise} and Eq. \eqref{eq-x0_t}, we get:
\begin{align}\label{eq-initial}  
R(k_1, 0) = x_0 - \hat{x}_{0|k_1} = \frac{\sqrt{1 - \bar{\alpha}_{k_1}}}{\sqrt{\bar{\alpha}_{k_1}}}( \epsilon_{\theta}(x_{k_1}, k_1) - \epsilon) 
\end{align}  

When transitioning from the step $k_i$ to a smaller step $k_j$, the residual propagation, 
based on Eq. \eqref{eq-ddim} and Eq. \eqref{eq-initial}, is defined as:
\begin{align}\label{eq-Rt2k} 
R(k_i,k_j) = \frac{\sqrt{1 - \bar{\alpha}_{k_j}}}{\sqrt{\bar{\alpha}_{k_j}}} (\epsilon_{\theta}(\hat{x}_{k_j}, k_j) -\epsilon_{\theta}(\hat{x}_{k_i}, k_i))
\end{align}
Where $\hat{x}_{k_j} = DDIM(\hat{x}_{0|{k_i}}, \epsilon_{\theta}(\hat{x}_{k_i}, k_i),k_j, 0)$, $\hat{x}_{k_1}=x_{k_1}$. 

\noindent \textbf{Path Residual: }
The cumulative residual along the entire reverse diffusion path (from
$k_1\xrightarrow{}\cdots \xrightarrow{}k_n\xrightarrow{}0$), 
is given by:
\begin{align}\label{eq-Rpath} 
R(k_n,0) = R(k_1,0) - \sum\limits_{i=1}^{n-1}R(k_i,k_{i+1})
\end{align}

Based on Eq. \eqref{eq-Rpath}, optimizing the path residual requires balancing the cumulative residuals along the path while also offsetting the initial residual. However, directly optimizing the path residual is not feasible due to the complexity of managing residuals across multiple steps. Instead, we focus on a target proxy that optimizes the initial residual. In the reverse process, transitioning from a later step to an earlier step refines the estimate of $x_0$ incrementally.
\textit{A smaller initial residual at $k_1$ leads to smaller residuals throughout the entire path, as each subsequent timestep builds upon this initial estimate}. Thus, optimizing the initial residual $R(k_1,0)$ can significantly reduce the cumulative residual, enhancing the overall traceability of the optimization.

\subsection{Optimization for Shortest-Path}
Building on the insights from Sec. \ref{RPPR}, directly optimizing cumulative residuals along the path brings significant challenges. However, if the optimal solution at an earlier step $k$ is known, it can be propagated backward to obtain the optimal solution at a later step $t$. This process enhances the initial residual, thereby optimizing the entire reverse diffusion path. To achieve this, we construct a \textit{step-reverse graph}, where edges are initialized between arbitrary steps $k$ and $t$ with $k < t$. An edge connecting $k$ to $t$, and shortest-path optimization is employed as a target proxy to minimize the cumulative residual error along the reverse diffusion path.

\noindent \textbf{Residual Elimination and Graph Construction:} 
However, directly defining edge weights based on residual propagation (e.g., $R(t,k)$) is not feasible. Ideally, the optimal solution at step $k$ should not be influenced by residuals at a later step $t$. Defining edge weights using residuals would introduce interference between steps, allowing optimal residual propagation to become untraceable.

% To resolve this, we use the true clean data $x_0$. Transitioning from $t$ to $k$ using $x_0$ via formula \eqref{eq-ddim} eliminates residuals along the path $t \xrightarrow{} k$, as $x_0$ provides an exact reference, bypassing intermediate residual propagation.
To address this issue, we begin with the clean data $x_0$ (due to it provides the precise reference) and then transition from $t$ to $k$ using $x_0$ via Eq.\eqref{eq-ddim} eliminates residuals along the path $t \xrightarrow{} k$. Thus, the residuals along $t \xrightarrow{} k$ are effectively counteracted.

We define the edge weight between step $k$ and $t$ as the residual difference computed using both the estimated $\hat{x}_0$ and the groundtruth $x_0$. This edge weight quantifies the error that accumulates when using the noisy estimate $\hat{x}_{0|t}$ compared to the true clean data $x_0$. 
Specifically, the edge weight is defined as follows:
\begin{align}\label{eq-edge} 
\text{edge}(k, t) = \lvert x_0 - \hat{x}'_{0|k} \rvert - \lvert x_0 - \hat{x}_{0|k} \rvert
\end{align}

Here, $\hat{x}'_{0|k}$ and $\hat{x}_{0|k}$ represent estimates of $x_0$ at step $k$, but from two different transformations. 
$\hat{x}'_{0|k}$ is estimated using $\hat{x}_{k}$, where:
\[
\hat{x}_{k} = \text{DDIM}(\hat{x}_{0|t}, \epsilon_{\theta}(x_{t}, t), k, 0)
\]
corresponds to the estimate-based transformation from timestep $t$ to $k$. Similarly, $\hat{x}_{0|k}$ is computed using $x_{k}$, where:
\[
x_{k} = \text{DDIM}(x_0, \epsilon_{\theta}(x_{t}, t), k, 0)
\]
represents the transformation based on the true clean data $x_0$. Both $\hat{x}'_{0|k}$ and $\hat{x}_{0|k}$ are then computed using the reverse process described in formula \eqref{eq-x0_t}.

By defining the edge weight in this manner, we ensure that the residuals are eliminated when using the true $x_0$ and that the propagation of residuals from an earlier step $k$ does not affect the optimal solution at a later step $t$. The $edge(k,t)$ represents the residual error from step $t$ to step $k$, illustrating the additional influence of the estimation error at step $t$ on the residual at step $k$ during the reverse diffusion process.

\noindent \textbf{Relaxation:} 
By utilizing the accurate clean data $x_0$, we eliminate the residuals along the diffusion path, ensuring that the result at step $k$ is optimal. With this optimal result at $k$, we can now employ shortest-path optimization to determine the optimal residual at step $t$.

% (i.e., the start timestep $k_1$).

The relaxation method is introduced to iteratively refine the residual at each step. Given the noisy data $x_t$, the objective is to minimize the residual error at step $t$, defined as \( dist(x_t, t) \). This is formulated as the absolute value of the initial residual $R(t, 0)$, i.e., $dist(x_t, t) = |R(t, 0)|=|x_0-\hat{x}_{0|t}|$.
% This also represents a near-optimal error among all possible paths, achieved through optimization. 
This represents a near-optimal error across all possible paths.
To ensure better optimization at step $t$, we propagate the optimal result from step $k$ 
% (i.e., \( t_n \)) 
using the following relaxation condition:
\begin{align}\label{eq-relax-cond}  
cond : dist(x_t, t) > dist(x_k, k) + edge(k, t)
\end{align}  
When this condition holds, the graph model chooses to transition directly from $k$ to $t$, as it provides a more optimal path wrt. residual error.
% Subsequently, the residual error at timestep $t$  is updated by propagating the optimal result from timestep $k$:
% \begin{align}\label{eq-update}  
% dist(x_t, t) \leftarrow dist(x_k, k) + edge(k, t)  
% \end{align}  
Consequently, the residual error at step $t$ is updated: $dist(x_t, t) \leftarrow dist(x_k, k) + edge(k, t)$. Through the relaxation and update process, we iteratively refine the path selection within the graph model, adjusting the residual at each step to minimize the residual error.

\subsection{Loss Function}

Based on Shortest-Path optimization, we design the total optimization loss as:
\begin{align}\label{eq-total-L}  
&L = \lambda \cdot L_{\epsilon} + cond \cdot L_{r},
\\
&L_{r} = \lVert dist(x_k, k) + edge(k, t) - dist(x_t, t) \rVert_2
\end{align}
where $\lambda$ is the noise loss weight, $cond$  the relaxation condition term from Eq.\eqref{eq-relax-cond}, and $L_r$ the relaxation loss,
% \begin{align}\label{eq-Lr}  
% L_{r} = \lVert dist(x_k, k) + edge(k, t) - dist(x_t, t) \rVert_2
% \end{align}
which allows the optimzal solution at step $k$ to guide the optimization at step $t$.
% The relaxation loss allows the optimal solution at timestep $k$ to guide the optimization at timestep $t$.
% Conceptually, this is equivalent to adding a reverse edge in the graph from $t$ to 0, with the edge weight defined as:
Inherently, this equals to adding a reverse edge in the graph from $t$ to 0, with the weight $edge(t, 0) = dist(x_k, k) + edge(k, t)$. 
% \[
% edge(t, 0) = dist(x_k, k) + edge(k, t).
% \]

Optimizing the model parameters at step $t$ effectively adds a new edge to the graph, connecting $t$ directly to $0$. This ``shortcut'' edge incorporates the contribution from the noisy input $x_t$.
Using the optimal solution at step $k$, the update reduces the residual at step $t$, thereby enhancing the reconstruction of the clean data $x_0$.
While graph-theoretic optimal solution is well-defined, the randomness and the ubiquitous presence of noise makes explicit graph construction and mathematical derivation of the optimal solution challenging. However, intuitively, network parameters can implicitly build the graph through learning, enabling us to approximate the optimal solution.

\begin{algorithm}[t]
\caption{Multi-State Optimization Training}
\label{alg:training}
\begin{algorithmic}[1]
\STATE Initialize $\epsilon_{\theta}$, $\epsilon_{\theta,\text{ema}}$, and \(\epsilon_{\theta,\text{graph}}\) with the same set of parameters.
\STATE Set moving average decay factor $\alpha$ for updating \(\epsilon_{\theta,\text{ema}}\).
Set iteration $N$ and interval $n$ for updating \(\epsilon_{\theta,\text{graph}}\).
\FOR {iteration $i = 1, 2, \dots, N$}
    \STATE Sample real data $x_0$ and a step $t$, then generate noisy data $x_t$ via Eq. \eqref{eq-addnoise}.
    \STATE Compute $L_{\epsilon}$ using $\epsilon_{\theta}$ via Eq. \eqref{eq-noiseloss}, and $dist(x_t,t) = |R(t,0)|$ using $\epsilon_{\theta}$ via Eq. \eqref{eq-x0_t},\eqref{eq-initial}.
    \STATE Randomly sample another step $k < t$. 
    Use $\epsilon_{\theta,\text{graph}}$ to obtain $\hat{x}_{0|t}$ via Eq. \eqref{eq-x0_t}, then transform $x_0$ and $\hat{x}_{0|t}$ to obtain $x_k$ and $\hat{x}_{k}$ using Eq. \eqref{eq-ddim}.
    \STATE Compute the edge weight $edge(k,t)$ using $\epsilon_{\theta,\text{graph}}$ via Eq. \eqref{eq-edge}.
    \STATE Use $\epsilon_{\theta,\text{ema}}$ to compute the reconstruction error $dist(x_k,k)$ at step $k$.
    \STATE Compute the relaxation condition  via Eq. \eqref{eq-relax-cond}.  
    % $cond = dist(x_t,t) > dist(x_k,k) + edge(k,t)$.
    % \STATE Compute the relaxation loss  via formula \eqref{eq-Lr}: 
    % $L_r = \|dist(x_k, k) + edge(k, t) - dist(x_t, t)\|_2$.
    \STATE Compute the total loss via Eq. \eqref{eq-total-L}: 
    % $L = \lambda L_{\epsilon} + cond \cdot L_r$.
    % \STATE Update $\epsilon_{\theta}$ and $\epsilon_{\theta,\text{ema}}$: 
    % $\theta \leftarrow \text{Optimize}(\theta, L), \theta_{\text{ema}} \leftarrow \alpha \theta_{\text{ema}} + (1 - \alpha) \theta$
    \STATE Update $\epsilon_{\theta}$ wrt. L, Update $\theta_{\text{ema}} \leftarrow \alpha \theta_{\text{ema}} + (1 - \alpha) \theta$
    
    \STATE Every $n$ iterations, 
    % copy $\epsilon_{\theta,\text{ema}}$ to $\epsilon_{\theta,\text{graph}}$:  
    $\theta_{\text{graph}} \leftarrow \theta_{\text{ema}}$.
\ENDFOR
\end{algorithmic}
\end{algorithm}

\subsection{Optimization Strategy}

To optimize the model effectively and stabilize the denoising process, as depicted in Algorithm \ref{alg:training}, we employ a multi-state optimization strategy that utilizes three instances of the model: the Base Model \(\epsilon_\theta\), the EMA Model \(\epsilon_{\theta,\text{ema}}\), and the Graph Model \(\epsilon_{\theta,\text{graph}}\). This strategy improves the stability of training in the presence of random noise.

The Base Model $\epsilon_\theta $ handles noise prediction and residual updates, serving as the fundamental logic for reconstruction.
The EMA Model $ \epsilon_{\theta,\text{ema}} $ smooths updates to stabilize training and provides more accurate estimates of the optimal error at timestep $k$. 
The Graph Model $ \epsilon_{\theta,\text{graph}} $ calculates edge weights for shortest-path optimization by employing delayed updates to capture global information and enhance stability in error accumulation. This combination of models ensures both stability and effectiveness in enhancing reconstruction quality within the shortest-path framework.

% The optimization process proceeds as follows: 
The whole optimization is unfolded as:
first, $ dist(x_t, t) $ is computed using $ \epsilon_\theta $, while $ dist(x_k, k) $ is computed using $ \epsilon_{\theta,\text{ema}} $, and the edge weights $ edge(k, t) $ are calculated using $ \epsilon_{\theta,\text{graph}} $. Then, the model parameters are updated based on the total loss function Eq.\ref{eq-total-L}.
% Periodically, 
At intervals the parameters of $ \epsilon_{\theta,\text{graph}} $ are synchronized to those of $ \epsilon_{\theta,\text{ema}} $ to maintain stability in edge weight calculations, ensuring accurate shortest-path optimization. This multi-state strategy not only improves robustness to random noise but also enhances the accuracy of error propagation across step, resulting in better denoising performance. 
It is worthy to note that, without graph modeling, ShortDF dengerates to DDIM.
% Detailed steps are provided in Algorithm \ref{alg:training}.

\subsection{Insights from Shortest Path Optimization}

% This section aims to demonstrate the effectiveness of the multi-state optimization strategy in achieving shortest path optimization. 
% By illustrating the method through a practical case study, we validate the theoretical foundations of the optimization approach and provide insights into its operational dynamics.
This section demonstrates how multi-state optimization helps achieve the shortest path in an effective manner. By illustrating a practical generation case, we validate the theoretical foundation of the proposed optimization strategy and provide insights into the operational mechanism.

% To illustrate the effectiveness of this optimization strategy, we present a case study focused on reducing the number of generation steps. 
We study making a case using a few generation steps.
For instance, prior to iterations, assume \( t = 10 \) and \( k = 2 \): paths \( 10 \to 0 \) and \( 10 \to 2 \to 0 \) represent one-step and two-step generation, respectively.

In iteration $1$, if the relaxation condition is satisfied, the reconstruction ability of path \( (10 \to 0) \) becomes equivalent to that of path \( (10 \to 2 \to 0) \), thus making one-step generation equivalent to the original two-step generation.

In iteration $2$, assuming \( t = 100 \) and \( k = 10 \), and the relaxation condition still holds, the path \( (100 \to 0) \) becomes equivalent to path \( (100 \to 10 \to 0) \). 
%Based on the result from iteration 1, 
Similarly, we can infer that path \( (100 \to 0) \) is equivalent to \( (100 \to 10 \to 2 \to 0) \), making one-step generation equivalent to the original three-step generation.
Continuing this reasoning through iteration \( n \), the path \( (T \to 0) \) becomes representative of the optimal error across various routes.

\section{Experiments}
\label{sec:exp}

\subsection{Settings}
\noindent \textbf{Datasets.} We evaluate our method on several widely adopted benchmarks, including CIFAR-10 ($32\times32$) \cite{Krizhevsky2009LearningML}, CelebA (CelebFaces Attributes Dataset) ($64\times64$) \cite{Liu2015Deep}, and LSUN Churches ($256\times256$) \cite{yu2015lsun}. The Fréchet Inception Distance (FID) \cite{Heusel2017gans} is utilized to assess image quality.

% Our code uses the official code and configuration of DDIM \cite{Song2021DDIM} and completes experimental training on a single NVIDIA GPU A100 80G. 
Our implementation utilizes the official code and configuration of DDIM \cite{Song2021DDIM} on a single NVIDIA A100 80GB GPU. We combine the designed shortest path loss with the original loss of DDIM to train the model. In the subsequent experiments conducted on three datasets, we employed the same noise input to generate DDIM samples and the output results of the proposed method, respectively. This approach explains the high degree of similarity observed in the generated samples.

\noindent \textbf{Models Compared} We compared with the most accelerated diffusion models available, \textit{e.g.} DDIM \cite{Song2021DDIM}, DPM-solver \cite{Lu2022DPM}, DPM-solver++ \cite{zheng2023dpm}, iPNDM \cite{Liu2022Pseudo}, DEIS \cite{Zhang2023Fast}, D-ODE \cite{kim2024distilling}, SDDPM \cite{cao2024spiking}, Resfusion \cite{shi2024resfusion}, and gDDIM \cite{Zhang2023gDDIM}.

\begin{table}[t]
  \centering
  \caption{Performance Metrics for CIFAR-10 on NVIDIA A100}
  \label{tab:cifar_results_merged}
  % Part A: Standard Evaluation
  \begin{small}
    \caption*{Part A: Standard Evaluation}
    \resizebox{0.48\textwidth}{!}{
    \begin{tabular}{l c c c c c}
      \toprule
      Method & Steps & FID ($\downarrow$) & Step Time (ms) & Speed-up & FID Improv. \\
      \midrule
      DDIM & 10 & 11.14 & 1.2 & 1.0x & - \\
      \midrule
      \textbf{ShortDF (Ours)} & 2 & \textbf{9.08} & 1.2 & \textbf{5.0x} & \textbf{+18.5\%} \\
      \bottomrule
    \end{tabular}
    }
  \end{small}

  \bigskip

  % Part B: Quantitative Comparisons
  \begin{small}
    \caption*{Part B: Quantitative comparisons across steps (NFE)}
    \resizebox{0.48\textwidth}{!}{
    \begin{tabular}{c c c c c c c c}
      \hline
      Method & 1 & 2 & 3 & 4 & 5 & 8 & 10 \\
      \hline
      DDIM & $>$100 & $>$100 & 123.54 & 66.13 & 26.91 & 19.06 & 11.14 \\
      DPM-solver & - & - & 90.38 & 33.29 & 23.31 & 6.91 & 5.09 \\
      DPM-solver++ & - & - & 107.02 & 30.46 & 18.87 & 12.58 & 7.83 \\
      iPNDM & - & - & - & $>$95 & 70.07 & $>$15 & 9.36 \\
      DEIS3 & - & - & - & $>$75 & 15.37 & $>$14 & 4.17 \\
      D-ODE & - & $>$100 & - & $>$40 & - & $>$10 & 8 \\
      SDDPM & - & - & - & 19.20 & - & 16.89 & - \\
      ResFusion & - & - & - & - & - & - & 28.82 \\
      gDDIM (DDPM) & - & - & - & - & - & - & 4.17 \\
      gDDIM (CLD) & - & - & - & - & - & - & 13.41 \\
      \hline
      \textbf{ShortDF (Ours)} & \textbf{25.07} & \textbf{9.08} & \textbf{7.31} & \textbf{6.59} & \textbf{6.45} & \textbf{4.93} & \textbf{3.75} \\
      \hline
    \end{tabular}
    }
  \end{small}
\end{table}

\subsection{Results}

\subsubsection{Quality Analysis on CIFAR-10 Dataset}

\begin{figure*}[t]
    \centering
    \includegraphics[width=0.78\linewidth]{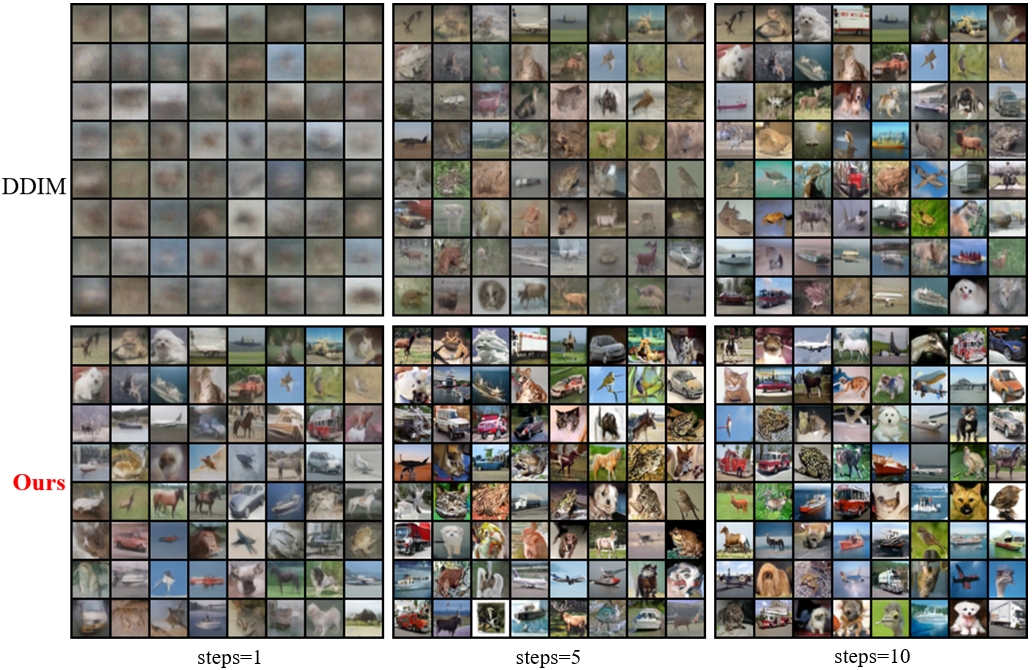}
    \caption{Sample images generated on the CIFAR-10 dataset at 1, 5, and 10 steps.}
    \label{fig_cifar}
\end{figure*}

To evaluate the effectiveness of the proposed method, we conducted a comprehensive analysis of the quality and speed of data generation using the CIFAR-10 dataset. The results are presented in Table \ref{tab:cifar_results_merged} and Figure \ref{fig_cifar}.

As shown in Table \ref{tab:cifar_results_merged}, our proposed ShortDF method achieves significant improvements in both speed and quality compared to the DDIM baseline. Specifically, in Part A of Table \ref{tab:cifar_results_merged}, ShortDF reduces the number of steps from 10 to 2 while achieving a lower FID score of 9.08, compared to DDIM's 11.14. This represents an 18.5\% improvement in FID while accelerating the generation process by 5.0x. These results demonstrate the efficiency and effectiveness of ShortDF in generating high-quality images with fewer steps.

In Part B of Table \ref{tab:cifar_results_merged}, we provide a detailed comparison across different steps (1 to 10). Our method consistently outperforms existing methods at each step level. For example, at 2 steps, ShortDF achieves an FID of 9.08, significantly lower than other methods, while maintaining fast generation speed. At 10 steps, ShortDF reaches an FID of 3.75, outperforming gDDIM (DDPM) which achieved an FID of 4.17. This indicates that ShortDF can generate high-quality images with fewer steps, making it highly efficient for low-resolution image generation tasks.

Figure \ref{fig_cifar} visually demonstrates the quality of images generated by ShortDF at different steps (1, 5, and 10). The results show that our method produces images with higher authenticity and clarity compared to DDIM, even with fewer steps. For instance, the images generated by ShortDF at 5 steps are comparable in quality to those generated by DDIM at 10 steps. This further highlights the ability of ShortDF to accelerate the diffusion model's data generation process while maintaining or improving image quality.

%In summary, our analysis shows that ShortDF significantly enhances the speed and quality of image generation on the CIFAR dataset. By reducing the number of steps required and achieving lower FID scores, ShortDF demonstrates its potential as an efficient and effective method for diffusion-based image generation.

\subsubsection{Quality analysis on CelebA dataset} 

\begin{figure*}[ht]
    \centering
    \includegraphics[width=0.85\linewidth]{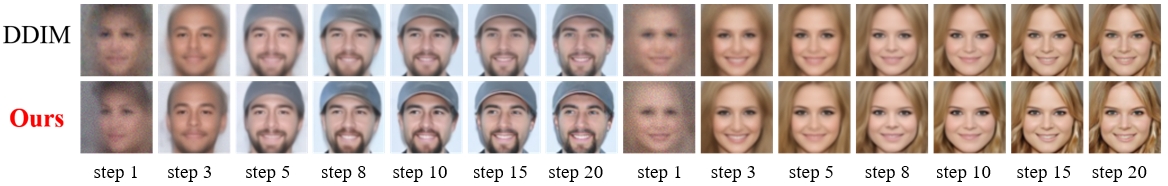}
    \caption{Performance comparison of our method and DDIM on the CelebA dataset at each time node along the same sampling path (with a total of 20 nodes), following the optimization of the initial residuals.}
    %\caption{The performance of our method with DDIM on the CelebA dataset at each time node along the same sampling path, following the optimization of the initial residuals.}
    \label{fig_CelebA}
\end{figure*}

On a dataset with a larger image size, CelebA, our method achieves excellent data generation results, as shown in Table \ref{CelebA}. Due to the larger image size of the CelebA dataset, the FID values under the 2 steps generation strategy are significantly higher than those observed in the CIFAR-10 dataset. The FID value of the proposed approach demonstrates a substantial reduction in the data creation strategy within 10 steps when compared to the baseline method, DDIM. Even with a 20 steps generation process, the FID value achieved by this research remains only half that of DDIM. The reduction in the FID value for the suggested method is comparable to that observed with DPM-Solver++. In contrast to this study, more advanced techniques such as iPNDM and DEIS exhibit slightly higher performance in the 10 steps generation strategy but show significantly greater values in other step configurations.

\begin{figure}[ht]
    \centering
    \includegraphics[width=0.90\linewidth]{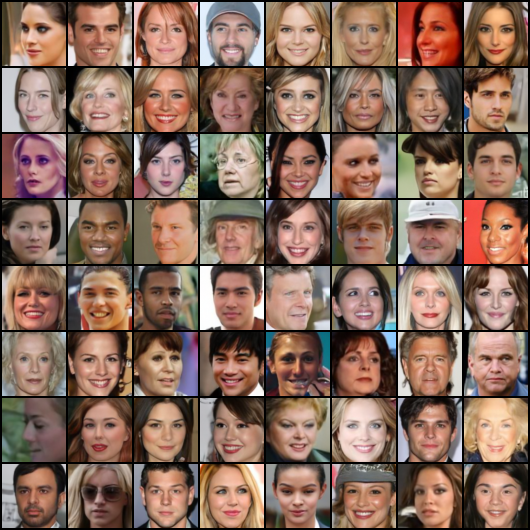}
    \caption{High-quality CelebA image samples generated using our method with only 10 diffusion steps.}
    %\caption{Samples on the CelebA dataset via 10 steps.}
    \label{fig_samp_celeba}
\end{figure}

The visualization results of several samples are presented in Fig. \ref{fig_CelebA} and Fig. \ref{fig_samp_celeba}. 

As shown in Fig. \ref{fig_CelebA}, the performance of our method and DDIM on the CelebA dataset is compared at each time node along the same sampling path. The figure clearly indicates that ShortDF exhibits much faster generation speed and higher clarity than DDIM when using the same step generation approach. Specifically, the image quality achieved with the 8th sampling (step 8) in this paper is quite comparable to that of DDIM's 15th and 20th samplings. This demonstrates the efficiency and effectiveness of our method in generating high-quality images with fewer steps.

Fig. \ref{fig_samp_celeba} presents more high-quality sampling instances generated by our method on the CelebA dataset via 10 steps. These samples further illustrate the superior performance of ShortDF in terms of image quality and generation speed.
% The visualization results generated from the data of several samples are presented in Fig.\ref{fig_CelebA} and Fig.\ref{fig_samp_celeba}. 
% A comparison of the findings in Fig.\ref{fig_CelebA} clearly indicates that the ShortDF exhibits much faster generation speed and clarity than DDIM when using the same step generation approach. The image quality achieved with the 8-step generation method in this paper is quite comparable to that of DDIM's 15-step and 20-step methods.
% %irrespective of the gender-specific generation results. 
% This document offers more comprehensive information and enhanced image quality produced using the 20-step technique.
\begin{table}[t]
\begin{center}
\begin{small}
\caption{Comparison results on the CelebA dataset within different steps (NFE).}
\label{CelebA}
\resizebox{1.0\linewidth}{!}{
\begin{tabular}{c c c c c c c c c c}
\hline
Method & 2 & 3 & 4 & 5 & 8 & 10 & 20 & 50 \\
\hline
DDIM & $>$100 & 57.65 & 38.01 & 27.20 & 13.94 & 10.59 & 6.35 &  4.66  \\
DPM-solver & - &  48.61 & 31.21 & 21.05 & 10.89 & 8.44 & 5.97 & 5.43 \\
DPM-solver++ & - &  61.26 & \textbf{19.75} & \textbf{14.70} & 10.24 & 8.31 & 5.94 & 5.40 \\
iPNDM & $>$100 & - & $>$70 & 59.87 & $>$20 & 7.78 & - & -   \\
DEIS3 & - & - & $>$40 & 25.07 & $>$15 & 6.95 & - & -  \\
D-ODE & $>$100 & - & $>$20 & - & $>$10 & $>$7.5 & - & -  \\
SDDPM & - & - & 25.09 & - & - & - & - & - \\
RDDM & - & - & - & - & - & 23.25 & - & -  \\
\hline
\textbf{ShortDF (Ours)} & \textbf{29.22} & \textbf{27.95} & 23.37 & \textbf{13.48} & \textbf{7.13} & \textbf{5.00} & \textbf{3.25} & \textbf{2.78} \\
\hline
\end{tabular}
}
\end{small}
\end{center}
\end{table}

\begin{table}[t]
\begin{center}
\caption{Comparison results on the Churches dataset within different steps (NFE).}
\begin{small}
\label{Churches}
\resizebox{1.0\linewidth}{!}{
\begin{tabular}{c c c c c c c c c c}
\hline
Method & 2  & 4 & 5 & 10 & 20 & 50 \\
\hline
DDIM & 168.72 & 122.62 & 69.53  & 21.52 & 8.24 & 7.65 \\
DPM-solver++ & - & \textbf{50.27} & 47.18 & 38.89 & 18.48 & 10.69  \\
PNDM & -  & - & \textbf{20.50} & 11.80 & 9.20 & 9.49  \\
\hline
\textbf{ShortDF (Ours)} & \textbf{134.91} & 66.48 & 40.38 & \textbf{11.78} & \textbf{7.97} & \textbf{6.95} \\
\hline
\end{tabular}
}
\end{small}
\end{center}
\end{table}

\begin{figure*}[ht]
    \centering
    \includegraphics[width=0.9\linewidth]{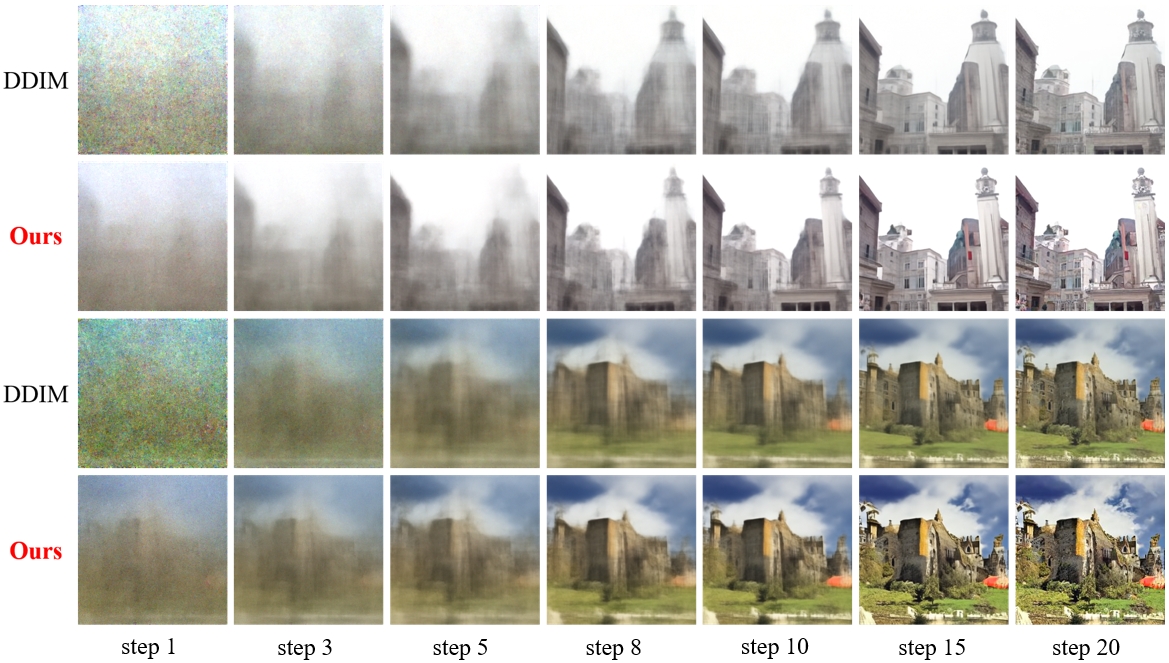}
    \caption{
    % Samples along a single path at varying timesteps on the Church dataset
    %Denoising quality as a function of varying timesteps, along a single path on the Church Dataset.
    Denoising quality of our method with DDIM on the Church dataset at each time node along the same sampling path, following the optimization of the initial residuals.
    }
    \label{fig_Church}
\end{figure*}

\begin{figure}[ht]
    \centering
    \includegraphics[width=0.90\linewidth]{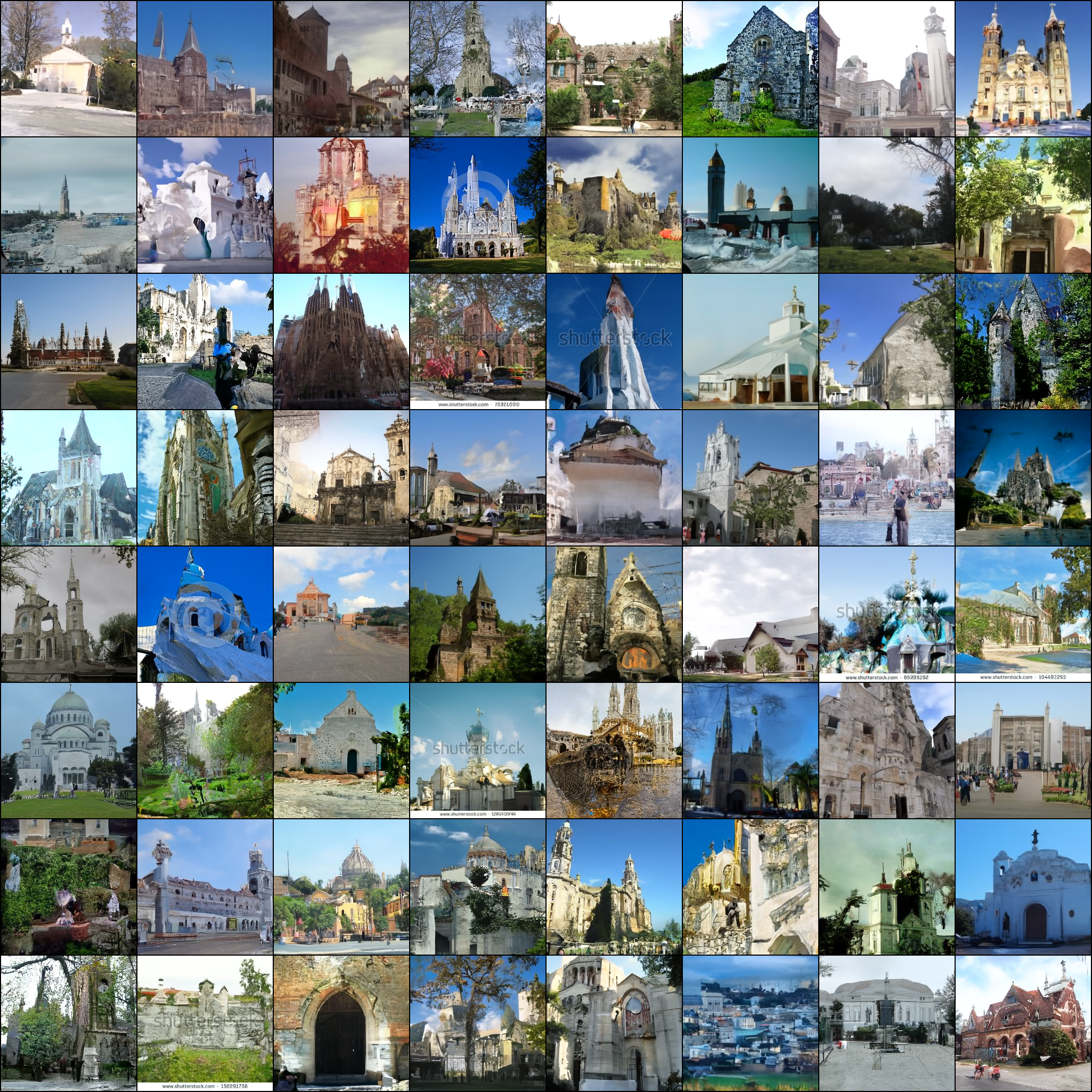}
    \caption{High-quality church image samples generated using our method with only 10 diffusion steps.}
    \label{fig_samp_church}
    \vspace{-5pt}
\end{figure}

\subsubsection{Quality analysis on Church dataset} 

The church dataset is a large collection of images that exceeds the sizes of both CIFAR-10 and CelebA. Its images often include additional elements such as sky and vegetation, which can significantly affect the quality of generated images. Consequently, there has been limited research conducted on this dataset. In this paper, we select the DDIM and PNDM methods for comparative analysis, as illustrated in Table \ref{Churches}. Due to the large image size, all methods yield relatively high FID values in fewer than five steps. When comparing the 2 steps and 4 steps strategies, our proposed method demonstrates a significantly lower FID value than DDIM and is slightly lower than DPM-Solver++ in the 5 steps strategy, although it performs better than the PNDM method. As the number of steps increases, the FID value decreases markedly, with values significantly lower than those of the PNDM method in the 10, 20, and 50 steps.

% The generated images align closely with the data presented in Fig. \ref{fig_samp_church}. 
As illustrated in Fig. \ref{fig_Church}, the DDIM approach exhibits increased noise at the first sampling node (step 1), rendering the object's contours nearly invisible. Additionally, the church casts a subtle shadow in our approach. The image quality of the proposed method at the 8th sampling node is remarkably similar to that of DDIM at the 15th sampling node. 
%However, when compared to DDIM's results with a 20-step strategy, the proposed method demonstrates a noticeable improvement under the 15-step strategy. 
It is important to highlight that the procedure from the 10th to the 20th sampling node employed in this article yields superior generation effects in the church background region, including the grass and blue sky.
% As demonstrated in Fig. \ref{fig_samp_church}, our method achieves high-quality church image generation with only 10 diffusion steps, closely aligning the generated samples with the target data.

Similarly, as demonstrated in Fig. \ref{fig_samp_church}, ShortDF achieves high-quality church image generation with only 10 diffusion steps, closely aligning the generated samples with the target data. This further highlights the efficiency and effectiveness of our approach in producing visually coherent results with minimal computational overhead.
\section{Conclusion}
This paper proposes a novel denoising diffusion method based on shortest path modeling, which enhances both the efficiency and quality of the denoising process by optimizing residual propagation. This approach significantly reduces sampling time while maintaining, and in some cases improving, the quality of the generated samples. To validate the method, we conduct extensive experiments on various benchmark datasets, demonstrating its state-of-the-art performance in addressing the computational overhead typically associated with diffusion models. This study demonstrates the effectiveness of incorporating graph theory into diffusion processes, providing valuable insights for future research and development. It not only advances the field of generative modeling but also establishes a solid foundation for real-time applications.

%\section{Acknowledgements}
%This work is supported by the National Natural Science Foundation of China (No. 62441610), the Natural Science Starting Project of SWPU (No. 2022QHZ023, 2022QHZ013), the Sichuan Science and Technology Program (No. 2024YFHZ0022), and the Sichuan Scientific Innovation Fund (No. 2022JDRC0009). 
%Due to double-blind review, the code and statement will be made public after the paper's acceptance.
%Due to double-blind review requirements, the code and statement for this paper will be made public after the paper has been officially accepted.
{
    \small
    \bibliographystyle{ieeenat_fullname}
    \bibliography{main}

\begin{thebibliography}{34}
\providecommand{\natexlab}[1]{#1}
\providecommand{\url}[1]{\texttt{#1}}
\expandafter\ifx\csname urlstyle\endcsname\relax
  \providecommand{\doi}[1]{doi: #1}\else
  \providecommand{\doi}{doi: \begingroup \urlstyle{rm}\Url}\fi

\bibitem[Austin et~al.(2021)Austin, Johnson, Ho, Tarlow, and van~den
  Berg]{Austin2021Discrete}
Jacob Austin, Daniel~D. Johnson, Jonathan Ho, Daniel Tarlow, and Rianne van~den
  Berg.
\newblock Structured denoising diffusion models in discrete state-spaces.
\newblock In \emph{NeurIPS}, pages 17981--17993, 2021.

\bibitem[Baranchuk et~al.(2022)Baranchuk, Voynov, Rubachev, Khrulkov, and
  Babenko]{Baranchuk2022Segmentation}
Dmitry Baranchuk, Andrey Voynov, Ivan Rubachev, Valentin Khrulkov, and Artem
  Babenko.
\newblock Label-efficient semantic segmentation with diffusion models.
\newblock In \emph{ICLR}, 2022.

\bibitem[Berthelot et~al.(2023)Berthelot, Autef, Lin, Yap, Zhai, Hu, Zheng,
  Talbott, and Gu]{Berthelot2023TRACT}
David Berthelot, Arnaud Autef, Jierui Lin, Dian~Ang Yap, Shuangfei Zhai, Siyuan
  Hu, Daniel Zheng, Walter Talbott, and Eric Gu.
\newblock {TRACT:} denoising diffusion models with transitive closure
  time-distillation.
\newblock \emph{CoRR}, abs/2303.04248, 2023.

\bibitem[Brempong et~al.(2022)Brempong, Kornblith, Chen, Parmar, Minderer, and
  Norouzi]{Brempong2022Denoising}
Emmanuel~Asiedu Brempong, Simon Kornblith, Ting Chen, Niki Parmar, Matthias
  Minderer, and Mohammad Norouzi.
\newblock Denoising pretraining for semantic segmentation.
\newblock In \emph{CVPRW}, pages 4174--4185, 2022.

\bibitem[Cao et~al.(2024)Cao, Wang, Guo, Cheng, Zhang, and Xu]{cao2024spiking}
Jiahang Cao, Ziqing Wang, Hanzhong Guo, Hao Cheng, Qiang Zhang, and Renjing Xu.
\newblock Spiking denoising diffusion probabilistic models.
\newblock In \emph{IEEE/CVF Winter Conference on Applications of Computer
  Vision}, pages 4912--4921, 2024.

\bibitem[Gu et~al.(2023)Gu, Zhai, Zhang, Liu, and Susskind]{Gu2023boot}
Jiatao Gu, Shuangfei Zhai, Yizhe Zhang, Lingjie Liu, and Josh Susskind.
\newblock Boot: Data-free distillation of denoising diffusion models with
  bootstrapping.
\newblock In \emph{Int. Conf. Mach. Learn.}, 2023.

\bibitem[Heusel et~al.(2017)Heusel, Ramsauer, Unterthiner, Nessler, and
  Hochreiter]{Heusel2017gans}
Martin Heusel, Hubert Ramsauer, Thomas Unterthiner, Bernhard Nessler, and Sepp
  Hochreiter.
\newblock Gans trained by a two time-scale update rule converge to a local nash
  equilibrium.
\newblock In \emph{NeurIPS}, 2017.

\bibitem[Ho et~al.(2020)Ho, Jain, and Abbeel]{Ho2020Denoising}
Jonathan Ho, Ajay Jain, and Pieter Abbeel.
\newblock Denoising diffusion probabilistic models.
\newblock In \emph{NeurIPS}, 2020.

\bibitem[Karras et~al.(2022)Karras, Aittala, Aila, and Laine]{Karras2022EDM}
Tero Karras, Miika Aittala, Timo Aila, and Samuli Laine.
\newblock Elucidating the design space of diffusion-based generative models.
\newblock In \emph{NeurIPS}, 2022.

\bibitem[Kim et~al.(2024)Kim, Tang, and Yu]{kim2024distilling}
Sanghwan Kim, Hao Tang, and Fisher Yu.
\newblock Distilling ode solvers of diffusion models into smaller steps.
\newblock In \emph{CVPR}, pages 9410--9419, 2024.

\bibitem[Krizhevsky(2009)]{Krizhevsky2009LearningML}
Alex Krizhevsky.
\newblock Learning multiple layers of features from tiny images.
\newblock Master's thesis, University of Toronto, 2009.

\bibitem[Li et~al.(2022)Li, Thickstun, Gulrajani, Liang, and
  Hashimoto]{Li2022Text}
Xiang~Lisa Li, John Thickstun, Ishaan Gulrajani, Percy Liang, and Tatsunori~B.
  Hashimoto.
\newblock Diffusion-lm improves controllable text generation.
\newblock In \emph{NeurIPS}, 2022.

\bibitem[Liu et~al.(2024{\natexlab{a}})Liu, Wang, Fan, Wang, Tang, and
  Qu]{Liu2024Residual}
Jiawei Liu, Qiang Wang, Huijie Fan, Yinong Wang, Yandong Tang, and Liangqiong
  Qu.
\newblock Residual denoising diffusion models.
\newblock In \emph{CVPR}, pages 2773--2783, 2024{\natexlab{a}}.

\bibitem[Liu et~al.(2022)Liu, Ren, Lin, and Zhao]{Liu2022Pseudo}
Luping Liu, Yi Ren, Zhijie Lin, and Zhou Zhao.
\newblock Pseudo numerical methods for diffusion models on manifolds.
\newblock In \emph{ICLR}, 2022.

\bibitem[Liu et~al.(2023)Liu, Gong, and Liu]{Liu2023Flow}
Xingchao Liu, Chengyue Gong, and Qiang Liu.
\newblock Flow straight and fast: Learning to generate and transfer data with
  rectified flow.
\newblock In \emph{ICLR}, 2023.

\bibitem[Liu et~al.(2024{\natexlab{b}})Liu, Zhang, Ma, Peng, and
  Liu]{Liu2024InstaFlow}
Xingchao Liu, Xiwen Zhang, Jianzhu Ma, Jian Peng, and Qiang Liu.
\newblock Instaflow: One step is enough for high-quality diffusion-based
  text-to-image generation.
\newblock In \emph{ICLR}, 2024{\natexlab{b}}.

\bibitem[Liu et~al.(2015)Liu, Luo, Wang, and Tang]{Liu2015Deep}
Ziwei Liu, Ping Luo, Xiaogang Wang, and Xiaoou Tang.
\newblock Deep learning face attributes in the wild.
\newblock In \emph{ICCV}, pages 3730--3738, 2015.

\bibitem[Lu et~al.(2022)Lu, Zhou, Bao, Chen, Li, and Zhu]{Lu2022DPM}
Cheng Lu, Yuhao Zhou, Fan Bao, Jianfei Chen, Chongxuan Li, and Jun Zhu.
\newblock Dpm-solver: {A} fast {ODE} solver for diffusion probabilistic model
  sampling in around 10 steps.
\newblock In \emph{NeurIPS}, 2022.

\bibitem[Meng et~al.(2023)Meng, Rombach, Gao, Kingma, Ermon, Ho, and
  Salimans]{Meng2023Distillation}
Chenlin Meng, Robin Rombach, Ruiqi Gao, Diederik~P. Kingma, Stefano Ermon,
  Jonathan Ho, and Tim Salimans.
\newblock On distillation of guided diffusion models.
\newblock In \emph{CVPR}, pages 14297--14306, 2023.

\bibitem[Salimans and Ho(2022)]{SalimansH2022Progressive}
Tim Salimans and Jonathan Ho.
\newblock Progressive distillation for fast sampling of diffusion models.
\newblock In \emph{ICLR}, 2022.

\bibitem[Shi et~al.(2024)Shi, Zheng, Xu, Dong, Pan, Xie, He, Li, and
  Fu]{shi2024resfusion}
Zhenning Shi, Haoshuai Zheng, Chen Xu, Changsheng Dong, Bin Pan, Xueshuo Xie,
  Along He, Tao Li, and Huazhu Fu.
\newblock Resfusion: Denoising diffusion probabilistic models for image
  restoration based on prior residual noise.
\newblock In \emph{NeurIPS}, 2024.

\bibitem[Song et~al.(2021)Song, Meng, and Ermon]{Song2021DDIM}
Jiaming Song, Chenlin Meng, and Stefano Ermon.
\newblock Denoising diffusion implicit models.
\newblock In \emph{ICLR}, 2021.

\bibitem[Song and Ermon(2020)]{Song2020Improved}
Yang Song and Stefano Ermon.
\newblock Improved techniques for training score-based generative models.
\newblock In \emph{NeurIPS}, 2020.

\bibitem[Song et~al.(2023)Song, Dhariwal, Chen, and
  Sutskever]{Song2023Consistency}
Yang Song, Prafulla Dhariwal, Mark Chen, and Ilya Sutskever.
\newblock Consistency models.
\newblock In \emph{Int. Conf. Mach. Learn.}, pages 32211--32252, 2023.

\bibitem[Wang et~al.(2023)Wang, Lu, Wang, Bao, Li, Su, and
  Zhu]{Wang2023ProlificDreamer}
Zhengyi Wang, Cheng Lu, Yikai Wang, Fan Bao, Chongxuan Li, Hang Su, and Jun
  Zhu.
\newblock Prolificdreamer: High-fidelity and diverse text-to-3d generation with
  variational score distillation.
\newblock In \emph{NeurIPS}, 2023.

\bibitem[Yang et~al.(2024)Yang, Zhang, Song, Hong, Xu, Zhao, Zhang, Cui, and
  Yang]{Yang2024Survey}
Ling Yang, Zhilong Zhang, Yang Song, Shenda Hong, Runsheng Xu, Yue Zhao, Wentao
  Zhang, Bin Cui, and Ming{-}Hsuan Yang.
\newblock Diffusion models: {A} comprehensive survey of methods and
  applications.
\newblock \emph{{ACM} Comput. Surv.}, 56\penalty0 (4):\penalty0 105:1--105:39,
  2024.

\bibitem[Yang and Mandt(2023)]{Yang2023Image}
Ruihan Yang and Stephan Mandt.
\newblock Lossy image compression with conditional diffusion models.
\newblock In \emph{NeurIPS}, 2023.

\bibitem[Yin et~al.(2023)Yin, Gharbi, Zhang, Shechtman, Durand, Freeman, and
  Park]{Yin2023One}
Tianwei Yin, Micha{\"{e}}l Gharbi, Richard Zhang, Eli Shechtman, Fr{\'{e}}do
  Durand, William~T. Freeman, and Taesung Park.
\newblock One-step diffusion with distribution matching distillation.
\newblock \emph{CoRR}, abs/2311.18828, 2023.

\bibitem[Yu et~al.(2015)Yu, Seff, Zhang, Song, Funkhouser, and
  Xiao]{yu2015lsun}
Fisher Yu, Ari Seff, Yinda Zhang, Shuran Song, Thomas Funkhouser, and Jianxiong
  Xiao.
\newblock Lsun: Construction of a large-scale image dataset using deep learning
  with humans in the loop.
\newblock \emph{arXiv preprint arXiv:1506.03365}, 2015.

\bibitem[Zhang and Chen(2023)]{Zhang2023Fast}
Qinsheng Zhang and Yongxin Chen.
\newblock Fast sampling of diffusion models with exponential integrator.
\newblock In \emph{ICLR}, 2023.

\bibitem[Zhang et~al.(2023)Zhang, Tao, and Chen]{Zhang2023gDDIM}
Qinsheng Zhang, Molei Tao, and Yongxin Chen.
\newblock gddim: Generalized denoising diffusion implicit models.
\newblock In \emph{ICLR}, 2023.

\bibitem[Zheng et~al.(2023{\natexlab{a}})Zheng, Nie, Vahdat, Azizzadenesheli,
  and Anandkumar]{Zheng2023Fast}
Hongkai Zheng, Weili Nie, Arash Vahdat, Kamyar Azizzadenesheli, and Anima
  Anandkumar.
\newblock Fast sampling of diffusion models via operator learning.
\newblock In \emph{ICLR}, pages 42390--42402, 2023{\natexlab{a}}.

\bibitem[Zheng et~al.(2023{\natexlab{b}})Zheng, Lu, Chen, and
  Zhu]{zheng2023dpm}
Kaiwen Zheng, Cheng Lu, Jianfei Chen, and Jun Zhu.
\newblock Dpm-solver-v3: Improved diffusion ode solver with empirical model
  statistics.
\newblock In \emph{NeurIPS}, pages 55502--55542, 2023{\natexlab{b}}.

\bibitem[Zhou et~al.(2024)Zhou, Chen, Wang, and Chen]{Zhou2023Fast}
Zhenyu Zhou, Defang Chen, Can Wang, and Chun Chen.
\newblock Fast ode-based sampling for diffusion models in around 5 steps.
\newblock In \emph{CVPR}, pages 7777--7786, 2024.

\end{thebibliography}
}


\end{document}